\documentclass[journal]{IEEEtran}
\usepackage{amsmath}
\usepackage{bm}
\usepackage{verbatim}
\usepackage{booktabs}
\usepackage{algorithm} 
\usepackage{algorithmic} 
\usepackage{multirow} 
\usepackage{xcolor}
\usepackage{caption}
\usepackage{graphicx}
\usepackage{subfigure}
\usepackage{float}
\usepackage{multirow}
\usepackage{multicol} 
\usepackage{colortbl,booktabs}
\ifCLASSINFOpdf
\else
\fi
\begin{document}
%
\title{Unsupervised Multi-modal Hashing for Cross-Modal Retrieval}
%
%
%

\author{Jun~Yu,
        Xiao-Jun~Wu*,
        Donglin Zhang
\thanks{J. Yu , X.-J. Wu (corresponding author) and D. Zhang are with the School of Artificial Intelligence and Computer Science, Jiangnan University, 214122, Wuxi, China. J. Yu, X.-J. Wu and D. Zhang are also with the Jiangsu Provincial Engineering Laboratory of Pattern Recognition and Computational Intelligence, Jiangnan University 214122, Wuxi, China.
e-mail: (yujunjason@aliyun.com;wu\_xiaojun@jiangnan.edu.cn;dlinzzhang@163.com).}
}

\maketitle

\begin{abstract}
With the advantage of low storage cost and high efficiency, hashing learning has received much attention in the domain of Big Data. In this paper, we propose a novel unsupervised hashing learning method to cope with this open problem to directly preserve the manifold structure by hashing. To address this problem, both the semantic correlation in textual space and the locally geometric structure in the visual space are explored simultaneously in our framework. Besides, the $\ell_{2,1}$-norm constraint is imposed on the projection matrices to learn the discriminative hash function for each modality. Extensive experiments are performed to evaluate the proposed method on the three publicly available datasets and the experimental results show that our method can achieve superior performance over the state-of-the-art methods.
\end{abstract}


%
\IEEEpeerreviewmaketitle

\section{Introduction}
\begin{figure*}[htbp]
\centering
\includegraphics[width=.8\textwidth]{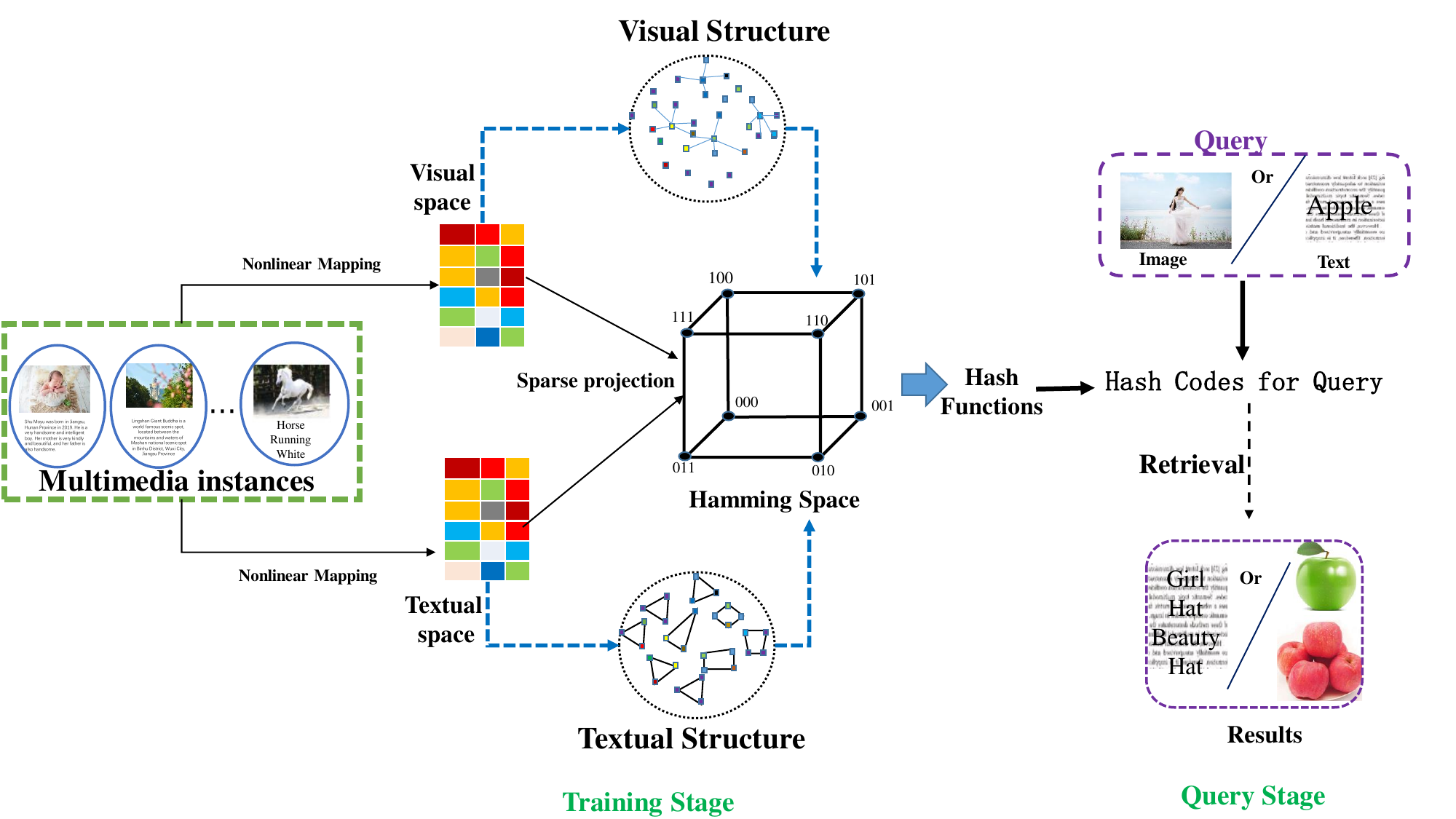}
\caption{ Illustration of the proposed approach. The proposed framework finds a discrete hamming space where the local geometric structure of visual space and the semantic correlation information provided by textual modality can be preserved simultaneously. In the query phase, we can obtain the hash codes of an arbitrary query according to the learned hash functions, and other modal data with the nearest hamming distance are returned. Best viewed in color.}
\label{Fig.1}
\end{figure*}

%
%
%
%
\IEEEPARstart{R}{ecently}, the explosive growth of multimedia data brings enormous challenge in information retrieval \cite{IEEETPAMI:Lin, IEEETIP:Guo}, data mining \cite{IEEETCSVT:Xiao, IEEETKDE:Wu}, and computer vision \cite{CVPR:Gong}. It is necessary to develop methods to support retrieving relevant objects from such massive database. Binary codes learning, a.k.a. hashing, has achieved great success because of its low storage and high efficiency. Among hashing methods \cite{IEEECVPR:Wang,IEEEICML:Norouzi,IEEE Transl:Yorozu,cvpr2019:li,tip2019:cui,tip2019:Lin,wmh:tang,IEEETIP:Ji}, Neighborhood Preserving Hashing (NPH)  \cite{cvpr2019:li}, Scalable Deep Hashing (SCADH) \cite{tip2019:cui}, Similarity Preserving Linkage Hashing (SPLH) \cite{tip2019:Lin}, Weakly Supervised Multimodal Hashing(WMH) \cite{wmh:tang} and Discrete Locally Linear embedding (DLLH) \cite{IEEETIP:Ji} have achieved promising performance. Nevertheless, these methods are assumed in single-modal circumstances and do not directly apply to multi-modal applications.\\
\hspace*{0.5cm}Cross-modal retrieval is a more interesting scenario because multimodal data are often available in multimedia domains. The major task of cross-modal retrieval is to find the same semantic data from different modal spaces when given query data. Most of the previous works pay attention to supervised and semi-supervised multimodal hashing learning algorithms that focus on learning discriminative features by ultilizing available semantic labels. Label Consistent Matrix Factorization Hashing (LCMFH) \cite{IEEETPAMI:Wang} learns a latent common space where data classified into the same category share a common representation. Multi-view Feature Discrete Hashing (MFDH) \cite{arxiv:Yu} jointly performs classifier learning and subspace learning for cross-modal retrieval. Semantic correlation maximization (SCM) \cite{AAAI:Zhang} reconstructs the semantic similarity matrix calculated by the label vectors in hamming space to learn the discriminative hash codes. Semantics-Preserving Hashing (SePH) \cite{IEEECVPR:Lin} transforms the semantic affinity into a probability distribution and approximates the distribution in Hamming space. Semi-supervised Hashing \cite{ICPR:Yu} learns the hash functions by utilizing the label information of partial data. Although the above methods are very efficient to realize cross-modal retrieval, they depend on the labeled data and it is time-consuming and labor-intensive to obtain them in real applications.\\
\hspace*{0.5cm}Unsupervised cross-modal hashing methods aim to learn the high-quality hash codes which preserve the structural and topological information of data. Cross-View Hashing  (CVH) \cite{AAAI:Kumar} is a pioneering work that extends the traditional unimodal spectral hashing \cite{NIPS:Weiss} to the multimodal situation. Robust Cross-view Hashing (RCH) \cite{IEEESPL:Shen} learns a common Hamming space in which the binary codes of the paired different modalities are as consistent as possible. Canonical Correlation Analysis (CCA) \cite{CVPR:Gong} transforms multiple views into a common latent subspace in which the correlation between two views is maximized. Fusion Similarity Hashing (FSH) \cite{CVPR:Liu} embeds the graph-based fusion similarity into a common Hamming space. The main idea of Inter-Media Hashing (IMH) \cite{ACM:Song} is that the learned binary codes preserve inter-media and intra-media consistency simultaneously. Unsupervised multimodal hashing generally needs to solve two basic problems: how to preserve the geometric structure among data points by hash codes and how to simultaneously select discriminative features for multiple modalities. Although existing unsupervised hashing methods have been developed, but above problems are not well addressed simultaneously. In fact, some tags or texts associated with uploaded images in social media contain the weakly semantic information. In this paper, we proposed a unsupervised multi-modal hashing where both the weakly semantic structure information provided by textual modality and the visually underlying manifold structure are explored simultaneously. Besides, the projection matrices are constrained by  $\ell_{2,1}$-norm to learn the discriminative and compact binary codes. The overview of the proposed method is shown in Fig.\ref{Fig.1} and the advantages of our method are summarized as follows\\
\hspace*{0.5cm}(1) We propose a sparse multi-modal hashing method by which the learned hash codes preserve the semantically and visually structural information.\\
\hspace*{0.5cm}(2)  Our model jointly performs the multi-modal graph embedding and discriminative features learning, which further improves the performance. \\
\hspace*{0.5cm}(2) Experimentally, a comparative evaluation of the proposed method on three available datasets with other state-of-the-art hashing methods shows that our method boosts the retrieval performance.\\
\hspace*{0.5cm}Structurally, the rest of this paper falls into four sections. In section \ref{sec2}, we simply introduce the related work in this field. Our model and the optimization algorithm are presented in section \ref{sec3}. In section \ref{sec4}, we discuss the experimental results on three available datasets and analyze the sensitivity of some parameters. Finally, the conclusions are drawn in section \ref{sec5}. 

\section{Related work}
\label{sec2}
In this section, we preliminarily review the related work in the field of cross-modal hashing. Cross-modal hashing algorithms are roughly divided into supervised cross-modal hashing and unsupervised cross-modal hashing which are distinguished by whether the label information is utilized or not.\\
\hspace*{0.3cm}Supervised cross-modal hashing methods learn the discriminative hashing feature via exploiting the available label information. Semantic Correlation Maximization (SCM) \cite{AAAI:Zhang} utilizes the semantic label to calculate the cosine similarity which is preserved in hamming space. Supervised Matrix Factorization Hashing (SMFH) \cite{rliu} integrates the graph regularization and matrix factorization into an overall hashing learning framework. Semantics-Preserving Hashing (SePH) \cite{IEEECVPR:Lin} transforms the affinity matrix into a probability distribution and approximates it in Hamming space via minimizing their Kullback-Leibler divergence. Generalized Semantic Preserving Hashing (GSePH) \cite{rMandal} preserves the semantic similarity by the unified binary codes. Semi-supervised NMF (CPSNMF) \cite{rwangcpsnmf} uses a constraint propagation approach to get more supervised information, which can greatly improve the retrieval performance. Cross-Modal Hamming Hashing (CMHH) \cite{rcao} designs a pairwise focal loss to generate compact and highly concentrated hash codes. In spite that supervised hashing methods have achieved promising performance, they overly depend on massive labeled data. Fortunately, unsupervised cross-modal hashing methods can handle effectively the problem.\\
\hspace*{0.3cm}Unsupervised cross-modal hashing methods mainly explore the structure, distribution, correlation and geometry among data and make these information be preserved well in hamming space. Canonical Correlation Analysis (CCA) \cite{CVPR:Gong} learns a common space where the correlation between different two modalities is maximized. Inter-Media Hashing (IMH) \cite{ACM:Song} introduces inter-media consistency and intra-media consistency to discover a common Hamming space. Cross View Hashing (CVH) \cite{AAAI:Kumar} extends the classical unimodal spectral hashing to the multi-modal scenario. Robust Cross-view Hashing(RCH) \cite{IEEESPL:Shen} learns a common Hamming space where the binary codes representing the same semantic content but different modalities should be as consistent as possible. Collective Reconstructive Embeddings (CRE) \cite{rhu} directly learns the unified binary codes via reconstructive embeddings collectively. Robust and Flexible Discrete Hashing(RFDH) \cite{rwang} adopts the discrete matrix decomposition to learn the binary codes, which avoids the large quantization error caused by relaxation. Fusion Similarity Hashing(FSH) \cite{CVPR:Liu} constructs an undirected asymmetric graph to model the similarity among objects.\\
\hspace*{0.3cm}Different from the above approaches, we propose a sparse multi-modal hashing approach that explores the local manifold structure and the wealkly semantic correlation to learn the robust hash functions. The $\ell_{2,1}$-norm regularization is incorporated to select the discriminative and relevant features from multi-modal data simultaneously.
\section{Unsupervised Multimodal Hashing}
\label{sec3}
\subsection{Notation and Problem Statement}
  Suppose that the training set $O=\{ o_i\}^{n}_{i=1}$ contains $n$ instances of image-text pair. $V = [v_1,v_2,...,v_n]\in{R^{d_1\times n}}$ and $T = [t_1,t_2,...,t_n]\in{R^{d_2\times n}}$ denote the image modality and text modality respectively. Each instance $o_i=(v_i,t_i)$ consists of an image $v_i\in{R^{d_1}}$ and a text $t_i\in{R^{d_2}}$. Without loss of generality, samples in each modality are zero-centered, i.e. $\sum_i^nv_i=0$ and $\sum_i^nt_i=0$. Given the code length $r$, all instance $O$ can be represented by the binary codes $B =[b_1;b_2;...b_n]\in{R^{n\times r}}$ in hamming space. We first calculate the kernel matrices $X^{(m)} = [x_1^{(m)},x_2^{(m)},...,x_n^{(m)}]\in R^{d\times n}(m=1,2)$ of the $m$-th modality by employing the RBF kernel function. Taking the image modality for an example,  $x^{(1)}_i=[exp(\|v_i-a_1\|^2/\sigma),...,exp(\|v_i-a_d\|^2/\sigma),]^T$, where $\{a_j\}_{j=1}^d$ are $d$ anchor points that are randomly selected from the image modality of the training data. The aims of our method is to learn the mapping functions from the kernel spaces to the common Hamming space, that is, $f:R^{d}\rightarrow{\{1,-1\}^r} $ for image-modality and $g:R^{d}\rightarrow{\{1,-1\}^r} $ for text-modality. \\
\hspace*{0.3cm}\textbf{Notations.} Given an example matrix $M$ and its $i-th$ row is $M_{i.}$, the $\ell_{2,1}$-norm of $M$ is defined as $\|M\|_{2,1}=\sum_{i=1}^n\sqrt{\sum_{j=1}^mM_{ij}^2}$.  $sgn(\cdot)$ signifies the sign function, specifically,
\begin{equation}
sgn(x)=
\begin{cases}
-1& x<0\\
\quad 1& x \ge 0
\end{cases}
\end{equation}

\subsection{Structure Preservation}
The local manifold structure in the original space should be preserved in the Hamming space. In the visual space, data point can be well approximated by the linear combination of its $k$-nearest neighbor points. In mult-imodal applications, the text content associated with an image can provide the weakly supervised semantic information. Thus, the semantic correlation of textual space should be considered in the process of the hashing learning. 
\subsubsection{Visual Model}
We hope that the similar locally manifold structure within visual modality can be projected into the same hash bin, and vice versa. The Discrete Locally Linear Embedding (DLLE) \cite{IEEETIP:Ji} is employed to preserve the local linear structure in the discrete Hamming space.  The reconstruction error is written as follows
\begin{equation}
\label{eq6}
\min_S\frac{1}{2}\sum_{i=1}^n\|x^{(1)}_i-\sum_{j\neq i}^nS_{ji}x^{(1)}_j\|^2
\end{equation}
where $S\in R^{n\times n}$ is an affinity matrix.The optimal solution can be obtained as follows
\begin{equation}
\label{eq7}
S_i = \frac{G_i^{-1}\textbf{1}}{\textbf{1}^TG_i^{-1}\textbf{1}}
\end{equation}
where  $X^{(1)}_j$ and $X^{(1)}_l$ are K-nearest neighbor points of $X^{(1)}_i$ and $G_i$ is the local Gram matrix defined as $(X^{(1)}_i-X^{(1)}_j)^T(X^{(1)}_i-X^{(1)}_l)$ for $X^{(1)}_i$. Each point can be approximated by an affine combination of its K nearest neighbors  in Hamming space. The reconstruction error as
\begin{equation}
\label{addeq1}
\min_B\|B-SB\|^2
\end{equation}
 
\subsubsection{Textual Model}
Texts associated with social images are usually provided by web users. These tagged texts with rich semantic information are beneficial to hash functions learning. The pairwise similarity matrix $Z \in R^{n \times n}$ is calculated using the cosine similarity function. The textual affinity between the $o_i$ and $o_j$ is defined as follows
\begin{equation}
\label{addeq1}
Z_{ij} = \frac{(x^{(2)}_i)^Tx^{(2)}_j}{\|x^{(2)}_i\|_2\|x^{(2)}_j\|_2}
\end{equation}
The higher textual similarity two instances have, the more similar binary codes they have. The above idea can be transformed the following weighted maximization problem
\begin{equation}
\label{addeq2}
\begin{split}
L(B,X^{(2)}) &= arg \max\limits_B \sum_{i,j=1}^n Z_{ij}b_ib_j^T\\& = arg \max\limits_B Tr(B^TZB)
\end{split}
\end{equation}
The  embedding scheme in Eq. (\ref{addeq2}) is termed Discrete Locally Projection Preservation (DLPP) in this paper. To achieve maximal information entropy, each hash bit is expected to be balanced on the training data \cite{NIPS:Weiss}. More specifically, the number of $+1$ and that of $-1$ should be consistent as much as possible for each bit. We integrate Eq. (\ref{addeq1}) and Eq. (\ref{addeq2}) into the following Eq. (\ref{eq8}) to obtain compact binary codes.
\begin{equation}
\label{eq8}
\begin{split}
\min_B&\|B-SB\|^2 - \beta Tr(B^TZB) + \rho\|\textbf{1}^T_nB\|_F^2
\\ &s.t.B\in \{-1,1\}^{n\times r}
\end{split}
\end{equation}
\subsection{Hash Functions Learning} 
The $\ell_{2,1}$-norm has been proven to be effective to obain the discriminative features by some recent
works \cite{IEEETPAMI:WangK,ZheChen:PR}. We impose the $\ell_{2,1}$-norm constraint on the projection matrices to learn the  discriminative representation, which leads to the following problem\\
\begin{equation}
\label{eq5}
\begin{split}
\min_{P^{(m)},B}\sum_{m=1}^2 &\alpha^{(m)^{\gamma}}(\|X^{(m)^T}P^{(m)}-B\|_F^2 + \lambda_m \|P^{(m)}\|_{2,1})\\& s.t.B\in \{-1,1\}^{n\times r},\sum_{m=1}^2\alpha_m=1,\alpha_m>0
\end{split}
\end{equation}
where $P^{(m)}$ denotes the projection matrix of the $m$-th modality and $\alpha^{(m)}$ is the weight factor with the adjustment coefficient $\gamma$ and $\lambda_m$ is a penalty parameter.

Then the overall objective function combining Eq. (\ref{eq8}) and Eq. (\ref{eq5}) is given as follows
\begin{equation}
\label{eq9}
\begin{split}
\min_{B,P^{(m)},\alpha^{(m)}}& \sum_{m=1}^2 \alpha^{(m)^{\gamma}}(\|X^{(m)^T}P^{(m)}-B\|_F^2 + \lambda_m \|P^{(m)}\|_{2,1})\\&
+\eta\|B-SB\|_F^2 - \beta Tr(B^TZB) + \rho\|\textbf{1}^T_nB\|_F^2 \\ & s.t.B\in \{-1,1\}^{n\times r},\sum_{m=1}^2\alpha_m=1,\alpha_m>0
\end{split}
\end{equation}
where $\beta$ and $\rho$ are two hyper-parameters.\\
\hspace*{0.3cm}The above Eq. (\ref{eq9}) is a non-convex problem. We solve the optimization problem by updating each variable with the other variables fixed alternatively.\\
\begin{bfseries}Update $\bm{B}$ with other variables fixed.\end{bfseries} The subproblem is to minimize the following 
\begin{equation}
\label{eq10}
\begin{split}
\min_B &\sum_{m=1}^2 \alpha^{(m)^{\gamma}}\|X^{(m)^T}P^{(m)}-B\|_F^2+\eta\|B-SB\|_F^2 \\&
- \beta Tr(B^TZB) + \rho\|\textbf{1}^T_nB\|_F^2 \\ &  s.t.B\in \{-1,1\}^{n\times r}
\end{split}
\end{equation}
The Eq.(\ref{eq10}) is an NP-hard problem since $B$ is constrainted to be discrete value. We relax it to be continuous value $H$. Thus the optimization problem can be transformed to
\begin{equation}
\label{eq11}
\begin{split}
\min_{H,B}&-2Tr(R^TH) + Tr(H^TH) + \eta\|CH\|_F^2  - \beta Tr(H^TZH) \\&+ \rho \|\textbf{1}^T_nH\|_F^2 + \xi \|H - B\|_F^2
\end{split}
\end{equation}
where $R=\sum_{m=1}^2\alpha^{(m)^\gamma}X^{(m)^T}P^{(m)}$ and $C=S-I$. Then we can get
\begin{equation}
\label{Heq1}
H = (\eta C^TC-\beta Z+\rho\textbf{1}_n\textbf{1}_n^T + (\xi + 1 )I)^{-1}(R + \xi B)
\end{equation}
The problem with respect to $B$ can be presented as
\begin{equation}
\label{Beq0}
\begin{split}
&\max_Btr(HB^T)\\& s.t. B\in \{-1,1\}^{n\times r}
\end{split}
\end{equation}
The soloution of $B$ can be directed obtained as 
\begin{equation}
\label{Beq1}
B = sgn(H)
\end{equation}

\begin{bfseries}Update $\bm{P^{(m)}}$ with other variables fixed.\end{bfseries} Keeping terms relating to $P^{(m)}$, the objective function Eq. (\ref{eq9}) can be rewritten as follows
\begin{equation}
\label{eq12}
\min_{P^{(m)}}\|X^{(m)^T}P^{(m)}-B\|^2+\lambda_m\|P^{(m)}\|_{2,1}
\end{equation}
Settting the derivative of Eq. (\ref{eq12}) with respect to $P^{(m)}$ to zero, we can obtain
\begin{equation}
\label{eq13}
P^{(m)} = (X^{(m)}X^{(m)^T}+\lambda_{(m)}D^{(m)})^{-1}X^{(m)}B
\end{equation}
where $D^{(m)}$ is a diagonal matrix with the $i$-th diagonal element $D^{(m)}_{ii}=\frac{1}{2\|P^{(m)}_i\|_2+\epsilon}$, and $P^{(m)}_i$ signifies the $i$-th row of $P^{(m)}$.

\begin{bfseries}Update weight $\bm{\alpha^{(m)}}$ with other variables fixed.\end{bfseries} By dropping terms irrelating to $\alpha^{(m)}$, we get
\begin{equation}
\label{eq14}
\begin{split}
&\min_{\alpha^{(m)}}\sum_{m=1}^2\alpha^{(m)^\gamma}C^{(m)} \\&s.t. \sum_{m=1}^2\alpha_m=1,\alpha_m>0
\end{split}
\end{equation}
where $C^{(m)}=\|X^{(m)^T}P^{(m)}-B\|_F^2 + \lambda_m \|P^{(m)}\|_{2,1}$. We employ the Lagrange multiplier to transform Eq. (\ref{eq14}) into the following
\begin{equation}
\label{eq15}
\min_{\alpha^{(m)}}\sum_{m=1}^2\alpha^{(m)^\gamma}C^{(m)}+\xi(1-\sum_{m=1}^2\alpha^{(m)})
\end{equation}
Setting the derivate of Eq. (\ref{eq15}) with respect to $\alpha^{(m)}$ to zero, we obtain
\begin{equation}
\label{eq16}
\alpha^{(m)}=\frac{(\gamma C^{(m)})^{1/(1-\gamma)}}{\sum_{m=1}^M(\gamma C^{(m)})^{1/(1-\gamma)}}
\end{equation}
After acquiring the projection matrix $P^{(m)}$, the binary codes $b$ of query $x$ is computed according to the rule $b = sgn(x^TP^{(m)})$. The overall optimization procedure is summarized in Algorithmn 1. This iteration process is repeated until it converges. As shown in Fig. \ref{Fig.4}, our algorithm converges quickly on the WiKi, PASCAL-VOC and UCI Handwritten Digit.
\renewcommand{\algorithmicrequire}{\textbf{Input:}} 
\renewcommand{\algorithmicensure}{\textbf{Output:}}
\begin{algorithm}
\caption{Unsupervised Multi-modal Hashing}
\label{Algorithm 1}
\begin{algorithmic}[1]
\REQUIRE $X^{(m)}\in R^{d\times n},(m=1,2)$; hash codes length $r$.
\ENSURE $P^{(m)}$, $B$, $\alpha^{(m)}$.\\
Initialize $B$, $P^{(m)}$, $\alpha^{(m)}$, $\lambda_{(m)}$, $\rho$, $\beta$ and $\eta$.\\
Compute similarity matrix $S$ according to (\ref{eq7}) \\
Compute similarity matrix $Z$ according to (\ref{addeq1})
\REPEAT
\STATE  Update $B$ according to (\ref{Heq1}) and (\ref{Beq1})\\
   \STATE  Compute $D^{(1)}_{ii}$ by $D^{(1)}_{ii}=\frac{1}{2\|P^{(1)}_i\|_2+\epsilon}$.\\
    \STATE  Compute $D^{(2)}_{ii}$ by $D^{(2)}_{ii}=\frac{1}{2\|P^{(2)}_i\|_2+\epsilon}$.\\
     \STATE Update $P^{(1)}$ using Eq.(\ref{eq13})
      \STATE Update $P^{(2)}$ using Eq.(\ref{eq13})
	 \STATE Update $\alpha^{(1)}$ according to Eq.(\ref{eq16})
	  \STATE Update $\alpha^{(2)}$ according to Eq.(\ref{eq16})
\UNTIL
\end{algorithmic}
\end{algorithm}

\section{Experiments}
\label{sec4}
\subsection{Datasets}
  \begin{bfseries}Wiki\end{bfseries} \cite{WiKi} contains 2,866 multimedia documents harvested from Wikipedia. Every document consists of an image and a text description, and each document is classified into one of  10 categories. Each image is represented by a 128-dimensional SIFT histogram vector. A 10-dimensional feature vector generated by latent Dirichlet allocation  is used to represent each text. We take 2173 pairs from the dataset to form the training set and database, the resting 973 as a query set.\\
 \hspace*{0.2cm} \begin{bfseries}PASCAL-VOC\end{bfseries} \cite{PASCAL-VOC} consists of  9,963 image-tag pairs. Each image is represented by a 512-dimensional GIST feature vector and each text is represented as a 399-dimensional word frequency count. Each sample are classified into one of 20 categories. We select 5,649 pairs with only one object in our experiment. 2,808 pairs are taken out as a training set and database, the remaining samples as the query data.\\
\hspace*{0.2cm} \begin{bfseries}UCI Handwritten Digit\end{bfseries} is comprised of handwritten numerals(0 - 9) collected from Dutch utility maps. Each of the character shapes is regarded as a class and each class consists of 200 samples. Following \cite{UCI}, we select 76 Fourier coefficients and 64 Karhunen-Love coefficients of the character shapes as the feature of two different modalities respectively. 1,500 samples are treated as the training set and database,  the resting 500 as the query set.
\subsection{Experimental Setting}
To verify the effectiveness of our method,  some comparative experiments are conducted on two cross-modal retrieval tasks: Image (Modality 1) query text (Modality 2) database and Text (Modality 2) query image (Modality 1) database which are termed as 'Task1' and 'Task2' respectively. As our method is a unsupervised hashing method, for a fair comparison, we compare our method with six state-of-the-art unsupervised learning models. Specifically, the baselines include CVH \cite{AAAI:Kumar}, CCA \cite{CVPR:Gong}, IMH \cite{ACM:Song}, RCH \cite{IEEESPL:Shen}, FSH \cite{CVPR:Liu} and CRE \cite{rhu}.  Since the source code of RCH and CRE is not available, we implemented it by ourselves. The codes of other baselines are kindly provided by the authors. The value of $\lambda_1$, $\lambda_2$, $\beta$ and $\rho$ are tuned in the candidate range $\{1e^{-5}, 1e^{-4},1e^{-3},1e^{-2},1e^{-1}, 1\}$. $\gamma$ is set to $0.5$ empirically and the best results are reported in this paper. Our experiments are implemented on MATLAB 2016b and Windows 10 (64-Bit) platform based on desktop machine with 12 GB memory and 4-core 3.6GHz CPU, and the model of the CPU is Intel(R) CORE(TM) i7-7700.

\begin{table}[h]
\centering
\caption{The mAP results on WiKi}
\label{Table 1}
\begin{tabular}{|c|c|c|c|c|c|}
\hline
\multirow{2}{*}{Tasks}&
\multirow{2}{*}{Methods}&
\multicolumn{4}{c|}{The length of hash code}\\
\cline{3 - 6}
&&16&32&64&128\\
\hline
\hline
\multirow{7}{*}{Task 1}&CVH&0.1499&0.1408&0.1372&0.1323\\
\cline{2-6}
&CCA&0.1699&0.1519&0.1495&0.1472\\
\cline{2-6}
&IMH&0.2022&0.2127&0.2164&0.2171\\
\cline{2-6}
&RCH&0.2102&0.2234&0.2397&0.2497\\
\cline{2-6}
&FSH&0.2346&0.2491&0.2531&0.2573\\
\cline{2-6}
&CRE&0.2301&0.2446&\bf{0.2599}&\bf{0.2620}\\
\cline{2-6}
&UMH&\bf{0.2511}&\bf{0.2505}&0.2578&0.2611\\
\hline
\multirow{7}{*}{Task 2}&CVH&0.1315&0.1171&0.1080&0.1093\\
\cline{2-6}
&CCA&0.1587&0.1392&0.1272&0.1211\\
\cline{2-6}
&IMH&0.1648&0.1703&0.1737&0.1720\\
\cline{2-6}
&RCH&0.2171&0.2497&0.2825&0.2973\\
\cline{2-6}
&FSH&0.2149&0.2241&0.2332&0.2368 \\
\cline{2-6}
&CRE&0.2442&0.2695&0.2846&0.2897\\
\cline{2-6}
&UMH&\bf{0.4984}&\bf{0.5057}&\bf{0.5224}&\bf{0.5298}\\
\hline

\multirow{7}{*}{Average}&CVH&0.1407&0.1290&0.1226&0.1208\\
\cline{2-6}
&CCA&0.1643&0.1456&0.1384&0.1341\\
\cline{2-6}
&IMH&0.1835&0.1915&0.1951&0.1946\\
\cline{2-6}
&RCH&0.2137&0.2365&0.2611&0.2735\\
\cline{2-6}
&FSH&0.2248&0.2366&0.2431&0.2470\\
\cline{2-6}
&CRE&0.2372&0.2571&0.2723&0.2759\\
\cline{2-6}
&UMH&\bf{0.3747}&\bf{0.3781}&\bf{0.3901}&\bf{0.3955}\\
\hline
\end{tabular}
\end{table}

\begin{table}[h]
\centering
\caption{The mAP results on PASCAL-VOC}
\label{Table 2}
\begin{tabular}{|c|c|c|c|c|c|}
\hline
\multirow{2}{*}{Tasks}&
\multirow{2}{*}{Methods}&
\multicolumn{4}{c|}{The length of hash code}\\
\cline{3 - 6}
&&16&32&64&128\\
\hline
\hline
\multirow{7}{*}{Task 1}&CVH&0.1484&0.1187&0.1651&0.1411\\
\cline{2-6}
&CCA&0.1245&0.1267&0.1230&0.1218\\
\cline{2-6}
&IMH&0.2087&0.2016&0.1873&0.1718\\
\cline{2-6}
&RCH&0.2633&0.3013&0.3209&0.3330\\
\cline{2-6}
&FSH&0.2890&0.3173&0.3340&0.3496\\
\cline{2-6}
&CRE&0.2758&0.3046&0.3216&0.3270\\
\cline{2-6}
&UMH&\bf{0.3225}&\bf{0.3368}&\bf{0.3741}&\bf{0.3701}\\
\hline
\multirow{7}{*}{Task 2}&CVH&0.0931&0.0945&0.0978&0.0918\\
\cline{2-6}
&CCA&0.1283&0.1362&0.1465&0.1553\\
\cline{2-6}
&IMH&0.1631&0.1558&0.1537&0.1464\\
\cline{2-6}
&RCH&0.2145&0.2656&0.3275&0.3983\\
\cline{2-6}
&FSH&0.2617&0.3030&0.3216&0.3428 \\
\cline{2-6}
&CRE&0.2395&0.2713&0.2941&0.2981\\
\cline{2-6}
&UMH&\bf{0.4760}&\bf{0.5472}&\bf{0.5825}&\bf{0.5701}\\
\hline

\multirow{7}{*}{Average}&CVH&0.1208&0.1066&0.1315&0.1165\\
\cline{2-6}
&CCA&0.1264&0.1315&0.1347&0.1386\\
\cline{2-6}
&IMH&0.1859&0.1787&0.1705&0.1591\\
\cline{2-6}
&RCH&0.2389&0.2834&0.3242&0.3657\\
\cline{2-6}
&FSH&0.2753&0.3102&0.3278&0.3462\\
\cline{2-6}
&CRE&0.2577&0.2880&0.3079&0.3126\\
\cline{2-6}
&UMH&\bf{0.3993}&\bf{0.4420}&\bf{0.4783}&\bf{0.4701}\\
\hline
\end{tabular}
\end{table}

\begin{table}[h]
\centering
\caption{The mAP results on UCI Handwritten Digit}
\label{Table 3}
\begin{tabular}{|c|c|c|c|c|c|}
\hline
\multirow{2}{*}{Tasks}&
\multirow{2}{*}{Methods}&
\multicolumn{4}{c|}{The length of hash code}\\
\cline{3 - 6}
&&16&32&64&128\\
\hline
\hline
\multirow{7}{*}{Task 1}&CVH&0.3421&0.2496&0.1907&0.1759\\
\cline{2-6}
&CCA&0.3155&0.2360&0.1841&0.2082\\
\cline{2-6}
&IMH&0.2947&0.2375&0.1892&0.1737\\
\cline{2-6}
&RCH&0.6181&0.6636&0.6991&0.7056\\
\cline{2-6}
&FSH&0.6323&0.6776&0.7027&0.7139\\
\cline{2-6}
&CRE&0.6636&0.7425&0.7516&0.7643\\
\cline{2-6}
&UMH&\bf{0.7496}&\bf{0.7944}&\bf{0.8149}&\bf{0.8043}\\
\hline
\multirow{7}{*}{Task 2}&CVH&0.3215&0.2471&0.1939&0.1695\\
\cline{2-6}
&CCA&0.3160&0.2398&0.1855&0.1102\\
\cline{2-6}
&IMH&0.2943&0.2315&0.1789&0.1514\\
\cline{2-6}
&RCH&0.5810&0.6336&0.6768&0.6979\\
\cline{2-6}
&FSH&0.6460&0.6745&0.7069&0.7149 \\
\cline{2-6}
&CRE&0.6448&0.7357&0.7547&0.7671\\
\cline{2-6}
&UMH&\bf{0.7327}&\bf{0.7997}&\bf{0.8333}&\bf{0.8417}\\
\hline

\multirow{7}{*}{Average}&CVH&0.3318&0.2483&0.1923&0.1727\\
\cline{2-6}
&CCA&0.3157&0.2379&0.1848&0.1592\\
\cline{2-6}
&IMH&0.2945&0.2345&0.1840&0.1626\\
\cline{2-6}
&RCH&0.5996&0.6486&0.6880&0.7017\\
\cline{2-6}
&FSH&0.6392&0.6761&0.7048&0.7144\\
\cline{2-6}
&CRE&0.6542&0.7391&0.7532&0.7657\\
\cline{2-6}
&UMH&\bf{0.7411}&\bf{0.7970}&\bf{0.8241}&\bf{0.8230}\\
\hline
\end{tabular}
\end{table}

\begin{figure*}[htbp]
\centering
\subfigure[]{
\includegraphics[width=.3\textwidth]{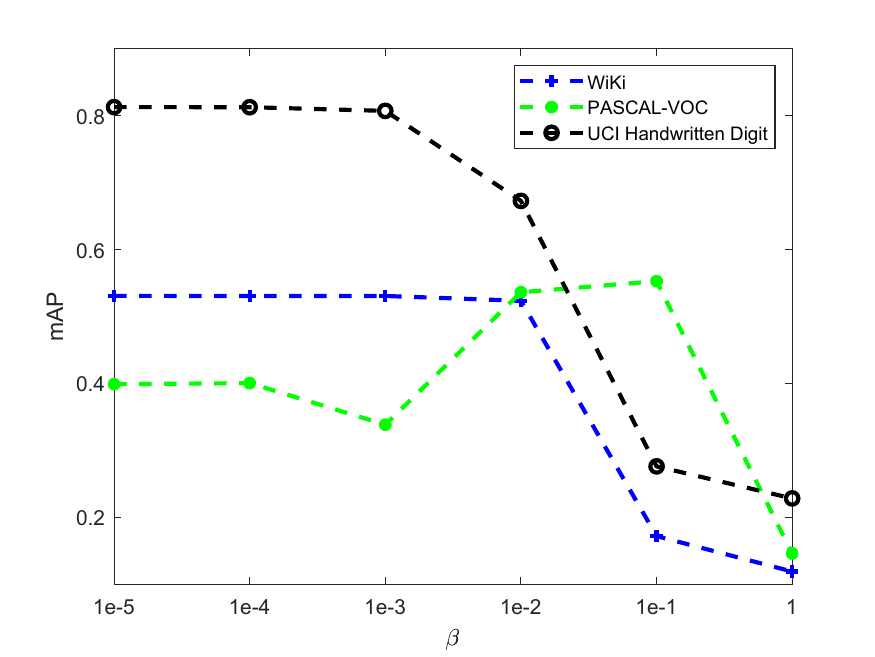}}
\subfigure[]{
\includegraphics[width=.3\textwidth]{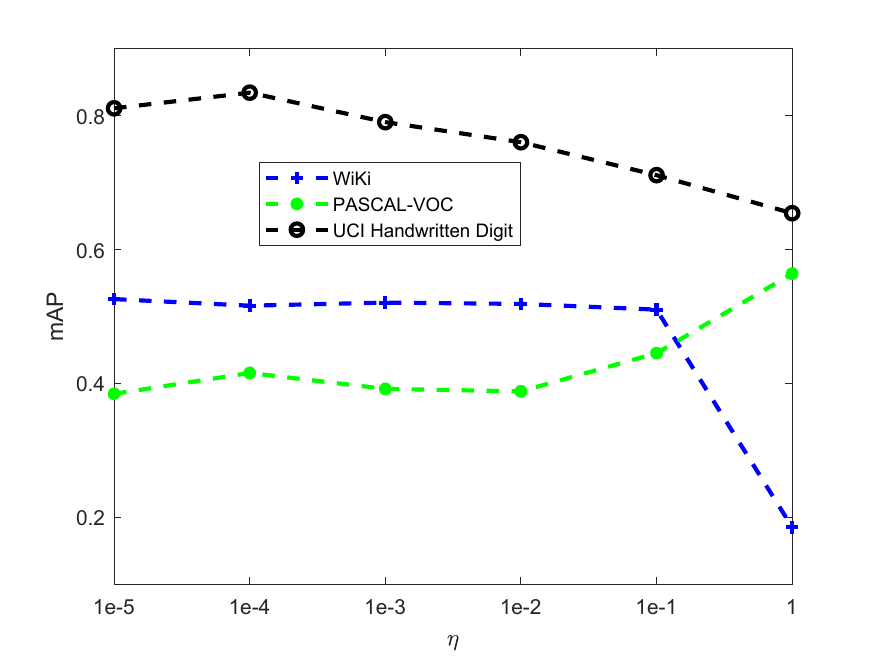}}
\subfigure[]{
\includegraphics[width=.3\textwidth]{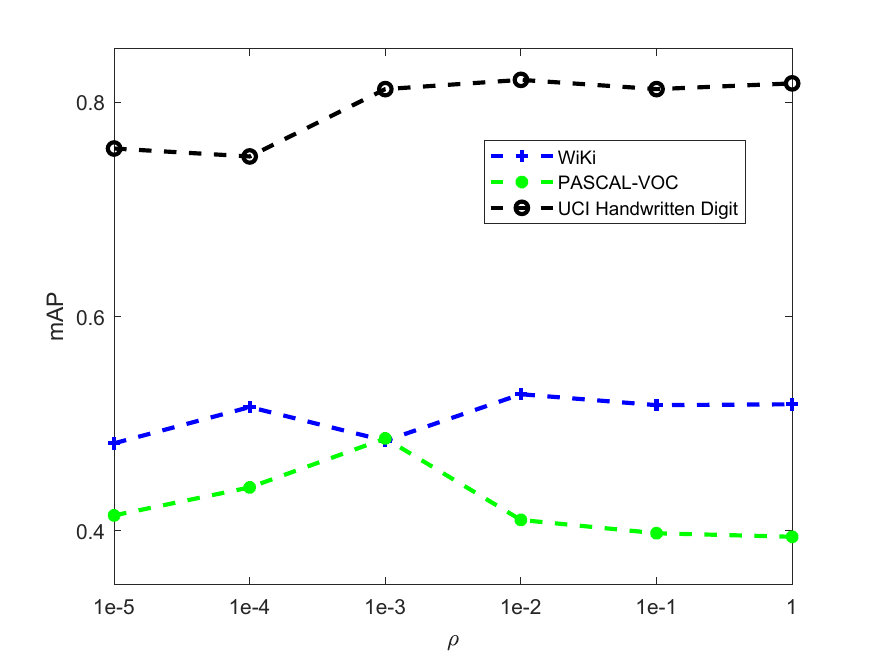}}
\caption{The mAP variation with respect to $\beta$, $\eta$ and $\rho$.}
\label{Fig.2}
\end{figure*}

\begin{figure*}[htbp]
\centering
\subfigure[WiKi]{
\includegraphics[width=.3\textwidth]{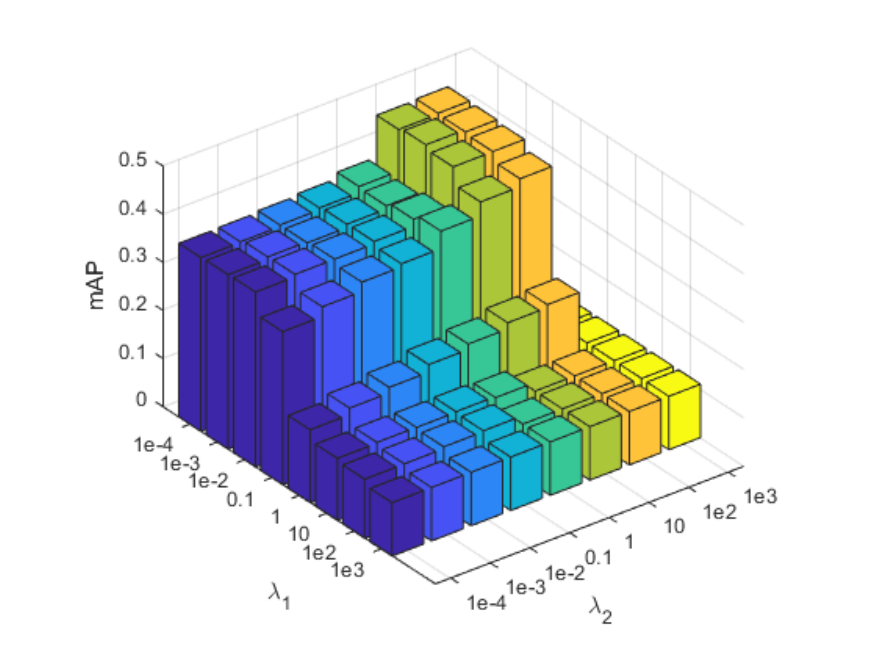}}
\subfigure[PASCAL-VOC]{
\includegraphics[width=.3\textwidth]{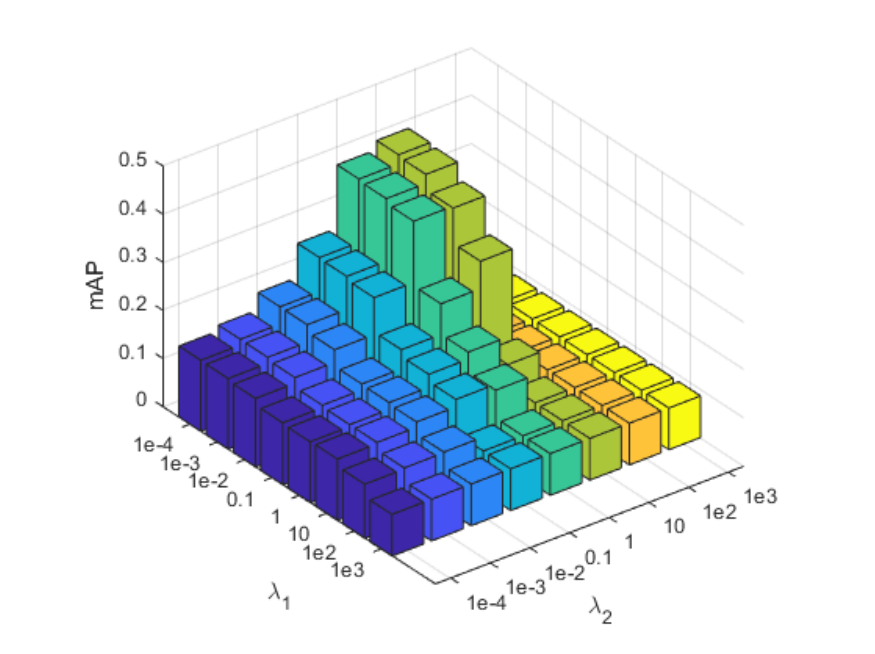}}
\subfigure[UCI Handwritten Digit]{
\includegraphics[width=.3\textwidth]{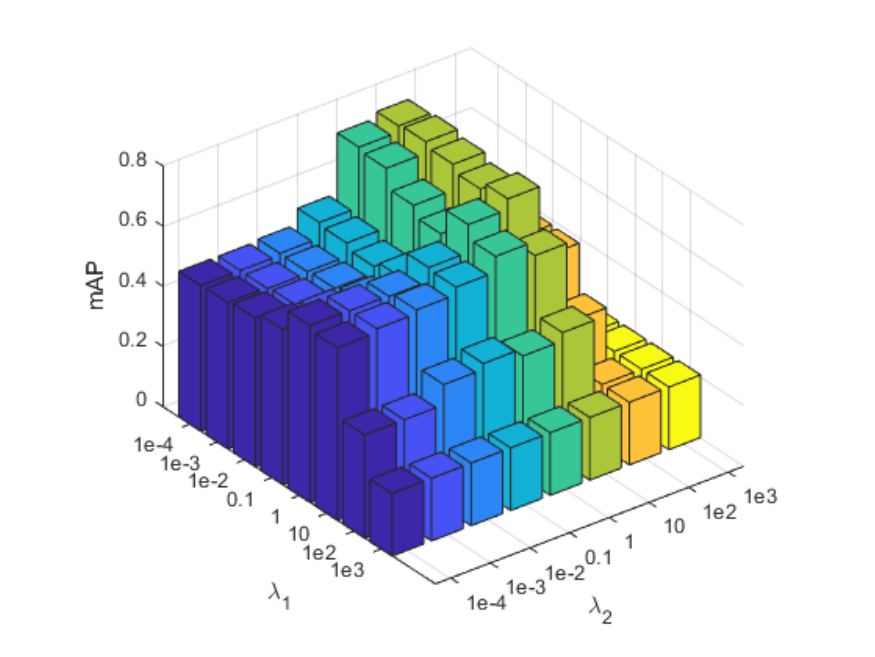}}
\caption{The mAP variation with respect to different combination of $\lambda_1$ and $\lambda_2$ on WiKi(a), PASCAL-VOC (b), and UCI Handwritten Digit (c).}
\label{Fig.3}
\end{figure*}

\begin{figure*}[htbp]
\centering
\subfigure[WiKi]{
\includegraphics[width=.3\textwidth]{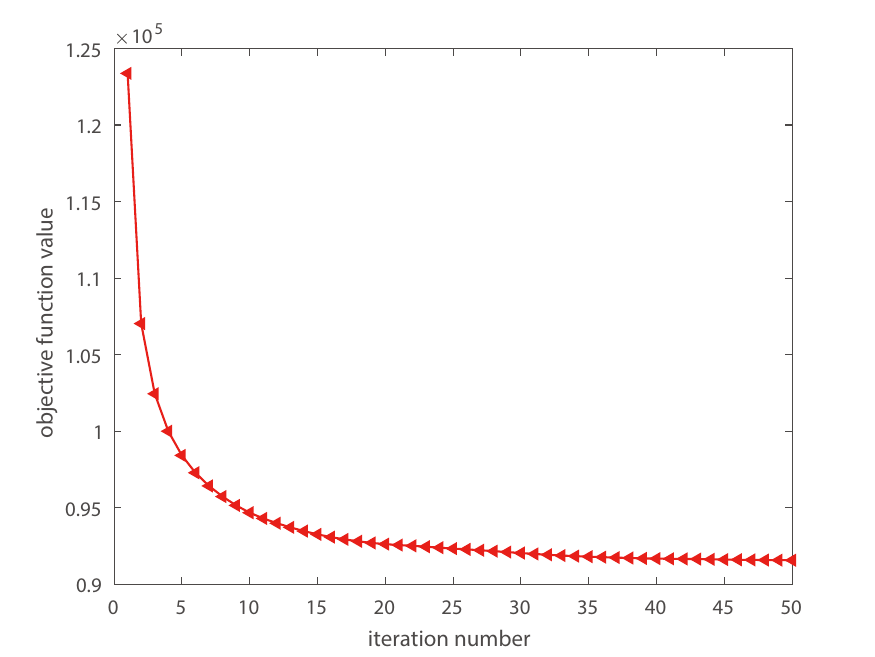}}
\subfigure[PASCAL-VOC]{
\includegraphics[width=.3\textwidth]{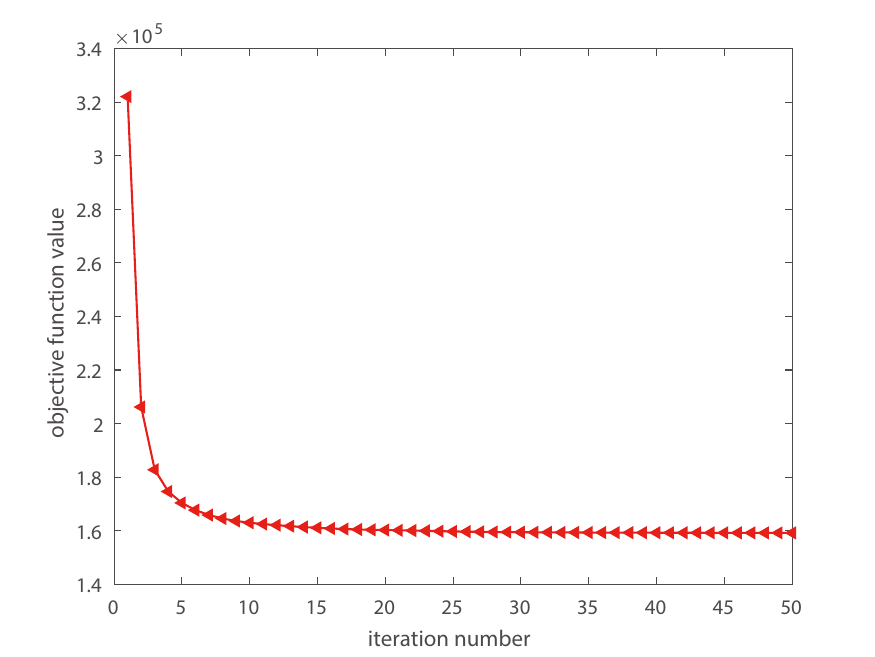}}
\subfigure[UCI Handwritten Digit]{
\includegraphics[width=.3\textwidth]{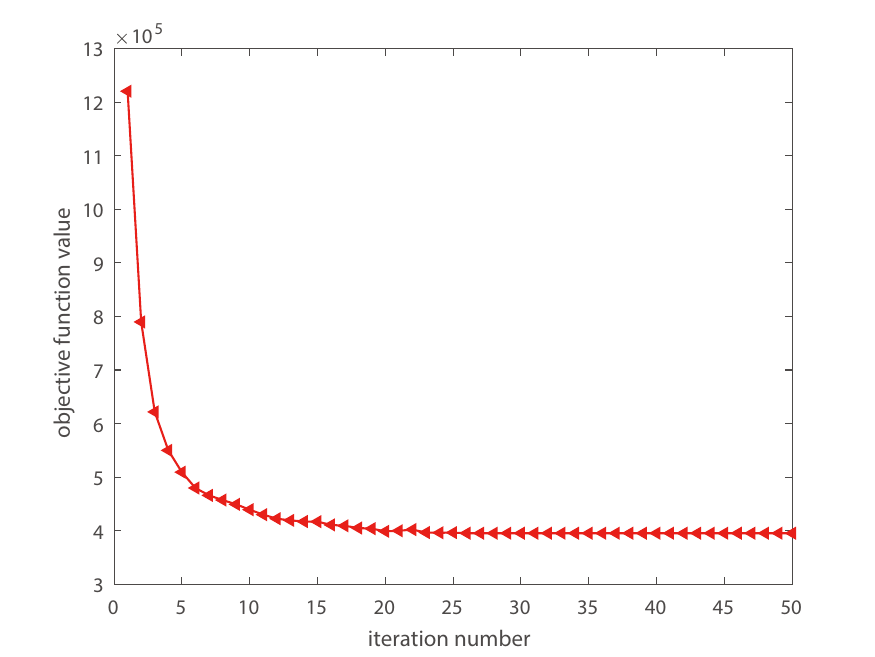}}
\caption{The convergence curve of algorithm 1 on WiKi(a), PASCAL-VOC (b), and UCI Handwritten Digit (c).}
\label{Fig.4}
\end{figure*}

\subsection{Evaluation metric}
The Mean Average Precision (mAP) is used to evaluate the performance of our method and comparison methods. Specifically, the Average Precision (AP) for a query $q$ is defined as follows
\begin{equation}
AP(q) =\frac{1}{l_q}\sum_{m=1}^RP_q(m)\delta_q(m)
\end{equation}
where $P_q(m)$ denotes the accuracy of top $m$ retrieval results; $\delta_q(m)=1$ if the $m$-th position is true neighbor of the query $q$, and otherwise  $\delta_q(m)=0$; $l_q$ is the correct statistics of top $R$ retrieval results. ($R$ is set to the number of entire database). The mAP is defined as the mean of the average precisions of all queries

\subsection{Retrieval Performance Evaluation}
The mAP scores on Wiki, PASCAL-VOC, and UCI Handwritten Digit are shown in TABLE \ref{Table 1}, \ref{Table 2}, and \ref{Table 3} respectively. We can observe the following points: (1) The performance of our method is superior to the baselines. Among baselines, RCH is a method with $\ell_{2,1}$-norm constraint imposing on projection matrices, but RCH does not take the manifold structure within each modality into account. FSH constructs an undirected asymmetric graph to model the similarity among samples. CRE utilizes domain-specific method to model different modalities and the intra-modal similarity is preserved in the process of learning unified binary codes.  However, FSH and CRE do not explore the discrimiantive features in their frameworks. Compared with the above methods,  our method boosts retrieval performance. The significant improvement of the proposed method can be attributed to the combination of the $\ell_{2,1}$-norm regularization and Structure Preserving. (2) Our method outperforms all comparison methods in terms of the average performance for two retrieval tasks on three datasets. With the increasing of hash code length, the retrieval performance on the Task 1 and Task 2 is further improved. The reason for the better performance is that the discriminative information will be more sufficient with the longer hash codes. (4) The results on Task 2 are consistently higher than that on Task 1. This may be because the text modality itself is a weak supervision information which can benefit to improve retrieval performance.

\begin{bfseries}Ablation study\end{bfseries} Some ablation experiments are conducted to investigate the influence of different terms in Eq. (9).  Three hyper-parameters ($\beta, \eta, \rho$) steer one of the terms respectively.  $\beta=0$ means our method ignores the inner structure within text modality. $\eta=0$ indicates our model does not consider the visually geometric information. $\rho=0$  implies the balanced bits term is ruled out from our model.  The comparison results are shown in Table \ref{table4}. From Table \ref{table4}, the importance of the three terms is dissimilar for different datasets. It is apparent that each term of the objective function collaboratively contributes to the retrieval performance.

\begin{table}[t]
\caption{Experiment results (mAP@64bit)  on ablation study}\smallskip
\centering
\begin{tabular}{ccccc}
\hline
&WiKi&Pascal VOC&UCI Handwritten Digit\\
\hline
$\beta = 0$&0.3801&0.3531&0.8014\\
$\eta = 0$&0.3850&0.3056&0.7998\\
$\rho = 0$&0.1102&0.4542&0.2082\\
Ours&\bf{0.3901}&\bf{0.4783}&\bf{0.8241}\\
\hline
\end{tabular}
\label{table4}
\end{table}

\subsection{Parameter Sensitivity Analysis}
In our model, $\rho$, $\beta$, $\eta$, $\lambda_1$ and  $\lambda_2$ are set manually. In this subsection, we explore the influence of different parameters setting on retrieval performance. The empirical analysis is performed for each parameter by varying its value in the candidate range. To discuss the above parameters conveniently, the hash code length is fixed at 64 bit in our experiments.  In Fig.\ref{Fig.2}, we plot the performance variation curves with respect to $\beta$, $\eta$ and $\rho$. On WiKi, PASCAL-VOC and UCI Handwritten Digit, our method can achieve the highest mAP score when $\beta$ is set to $1e^{-5}$, $1e^{-1}$ and $1e^{-6}$  respectively and $\eta$ is set to $1e^{-1}$, $1$ and $1e^{-4}$  respectively. $\rho$ represents the importance of the balanced term in Eq. (\ref{eq9}). In Fig.\ref{Fig.2}, we can find that the $\rho$ should not be too large. $\lambda_1$ and $\lambda_2$ are two penalty parameters controlling the sparse constraint items of two modalities respectively. The mAP scores as a function of $\lambda_1$ and $\lambda_2$ is plotted in  Fig.\ref{Fig.3}, which shows that the optimal combination falls a fixed small range on three datasets.

\section{Conclusion}
\label{sec5}
In this paper, we propose a unsupervised multi-modal hashing method for cross-modal retrieval. Our model explores the underlying neighborhood structure of the visual space and the semantic correlation provied by textual modality to learn the compact unified hash codes. The sparse constraint is imposed on our model to learn discriminative hash functions for multi-modal data. Encouraging experimental results demonstrate that the effectiveness of the proposed framework on cross-modal retrieval tasks. In the future, we plan to extend the proposed method into the deep learning networks.

\section*{Acknowledgment}
THE PAPER IS SUPPORTED BY THE NATIONAL NATURAL SCIENCE FOUNDATION OF CHINA(GRANT NO.61672265, U1836218), AND THE 111 PROJECT OF MINISTRY OF EDUCATION OF CHINA (GRANT NO. B12018).

\ifCLASSOPTIONcaptionsoff
  \newpage
\fi

\end{document}